\documentclass[10pt,twocolumn,letterpaper]{article}

\usepackage{iccv}
\usepackage{times}
\usepackage{epsfig}
\usepackage{graphicx}
\usepackage{amsmath}
\usepackage{amssymb}

\usepackage{graphicx}
\usepackage{amsmath}
\usepackage{amssymb}
\usepackage{booktabs}
\usepackage{multirow}
\usepackage[accsupp]{axessibility}

\usepackage[pagebackref=false,breaklinks=true,letterpaper=true,colorlinks,bookmarks=false]{hyperref}

\iccvfinalcopy %
\usepackage[capitalize]{cleveref}
\crefname{section}{Sec.}{Secs.}
\Crefname{section}{Section}{Sections}
\Crefname{table}{Table}{Tables}
\crefname{table}{Tab.}{Tabs.}

\ificcvfinal\fi

\begin{document}

\title{LongSplat: Online Generalizable 3D Gaussian Splatting from Long Sequence Images}

\author{
  Guichen Huang$^{1,2}$,
  Ruoyu Wang$^{3}$,
  Xiangjun Gao$^{4}$,
  Che Sun$^{2}$,
  Yuwei Wu$^{2,1}$\thanks{Corresponding author.},
  Shenghua Gao$^{3,5}$,
  Yunde Jia$^{2,1}$
  \and
  \textnormal{$^1$Beijing Institute of Technology  \quad $^2$ Shenzhen MSU-BIT University \quad $^3$ Transcengram}
  \and
  \textnormal{
   $^4$ The Hong Kong University of Science and Technology \quad $^5$ The University of Hong Kong
  }
}

\maketitle
\ificcvfinal\thispagestyle{empty}\fi

\begin{abstract}
3D Gaussian Splatting achieves high-fidelity novel view synthesis, but its application to online long-sequence scenarios is still limited. 
Existing methods either rely on slow per-scene optimization or fail to provide efficient incremental updates, hindering continuous performance.
In this paper, we propose LongSplat, an online real-time 3D Gaussian reconstruction framework designed for long-sequence image input. 
The core idea is a streaming update mechanism that incrementally integrates current-view observations while selectively compressing redundant historical Gaussians. 
Crucial to this mechanism is our Gaussian-Image Representation (GIR), a representation that encodes 3D Gaussian parameters into a structured, image-like 2D format.
GIR simultaneously enables efficient fusion of current-view and historical Gaussians and identity-aware redundancy compression. 
These functions enable online reconstruction and adapt the model to long sequences without overwhelming memory or computational costs.
Furthermore, we leverage an existing image compression method to guide the generation of more compact and higher-quality 3D Gaussians.
Extensive evaluations demonstrate that LongSplat achieves state-of-the-art efficiency-quality trade-offs in real-time novel view synthesis, delivering real-time reconstruction while reducing Gaussian counts by 44\% compared to existing per-pixel Gaussian prediction methods. 




\end{abstract}

\section{Introduction}

Growing interest in 3D scene reconstruction and novel view synthesis has led to rapid advancements in the field, among which 3D Gaussian splatting(3DGS) \cite{kerbl20233d,yu2024mip,huang20242d,lu2024scaffold} has gained particular attention for its effectiveness. 
Despite its impressive rendering speed at inference time, most existing methods still rely on slow, per-scene optimization for reconstruction, which can take minutes to hours even for moderately sized environments. 
This slow optimization is a significant barrier for applications requiring fast, real-time perception and response, such as embodied AI and robotics, where timely adaptation to dynamic environments is essential. 
To address these challenges, there is an increasing need for systems that can process long sequences of visual data in real-time, dynamically updating with each new frame input while ensuring high-quality reconstruction.

Recent efforts have aimed to improve reconstruction efficiency by developing generalizable splatting models that directly predict 3D Gaussian parameters from images in a feed-forward manner.
These methods \cite{charatan2024pixelsplat,xu2024depthsplat,li2025streamgs} significantly reduce processing time and perform well under sparse-view settings.
However, their performance often degrades when applied to long sequences or dense multi-view scenarios: the reconstructed Gaussians become increasingly redundant and noisy, resulting in artifacts such as floating points and blurred regions.
Moreover, memory and computational costs grow rapidly as more views are processed, making these approaches difficult to scale to real-world applications involving hundreds of frames.
These limitations arise primarily from two factors: a lack of global historical Gaussians modeling and the absence of an efficient incremental update mechanism, both of which are essential for robust long-term reconstruction.
Although some recent works \cite{wang2025zpressor,wang2024freesplat,wang2025freesplat++,ziwen2024long, li2025streamgs} extending generalizable 3D GS to sequential inputs sets, they still struggle with incremental updates or rely on fixed-length reconstruction pipelines, limiting their flexibility and scalability in online long-sequence scenarios.

In this paper, we propose LongSplat, an online 3DGS framework designed for real-time, incremental reconstruction from long-sequence images. 
Its core innovation lies in an incremental update mechanism that integrates current-view observations while selectively compressing redundant historical Gaussian.
This mechanism efficiently performs two key operations per frame: 
(1) Adaptive Compression: selectively compressing accumulated Gaussians from past views to eliminate redundancy and minimize storage/rendering costs, 
and (2) Online Integration: fusing current-view Gaussians with the historical state.
These strategies aim to mitigate a core limitation of generalizable 3DGS: per-pixel prediction inherently produces dense but redundant Gaussians.
By progressively refining the Gaussian field over time, our method seeks to improve scalability and memory usage while enhancing consistency across views.
In addition, the compression mechanism reduces redundancy and offers a potential path toward dynamic scene modeling, where outdated or redundant elements can be removed in a lightweight, incremental manner without reprocessing the entire sequence.

Specifically, we propose Gaussian-Image Representation (GIR) that projects 3D Gaussian parameters into a structured 2D image-like format. 
This representation enables online reconstruction by facilitating the propagation of information across views and supporting localized compression.
To enhance cross-view interaction, GIR projects historical Gaussians into the current frame, enabling feature-level fusion.
This fusion not only improves the spatial consistency of the reconstructed 3D Gaussian field, but also provides a structured basis for subsequent compression of redundant historical information.
In addition, GIR plays a central role in localized compression by maintaining the mapping between 2D projections and their corresponding historical 3D Gaussians. 
This identity-aware structure makes 3DGS more tractable and removes redundant splats accumulated over time. 
Such compression not only reduces memory and rendering cost, but also improves visual quality by eliminating overlapping or outdated Gaussians.
Furthermore, we leverage GIR's image-like structure to apply supervision from ground-truth 3DGS, using an optimized per-scene Gaussian dataset constructed with existing image compression techniques \cite{fan2024lightgaussian}.
This strategy improves both compactness and fidelity of the learned 3D Gaussians without requiring full 3D loss computation.

Through extensive evaluations, we demonstrate that LongSplat achieves state-of-the-art efficiency-quality trade-offs in real-time novel view synthesis. 
Our method achieves real-time rendering and reduces Gaussian counts by 44\% on DL3DV\cite{ling2024dl3dv}. 
Moreover, LongSplat outperforms the baseline by 3.6 dB in PSNR on the DL3DV benchmark and exhibits superior scalability for long-sequence scene reconstruction. By efficiently processing long-sequence visual data, LongSplat opens up new possibilities for real-time 3D perception in applications that require handling extensive visual inputs.
Our contributions can be summarized as follows:
\begin{itemize}
    \item We propose LongSplat, a real-time 3D Gaussian reconstruction framework tailored for arbitrary-view, long-sequence image inputs. By introducing a 3D Gaussian updating mechanism that selectively compresses redundant historical Gaussians and incrementally integrates current-view observations, LongSplat enables scalable, memory-efficient reconstruction and real-time novel view synthesis.
    \item We introduce Gaussian-Image Representation(GIR), a structured 2D representation of 3D Gaussians that enables efficient historical feature fusion, redundancy compression, lightweight 2D operations, and GIR-space supervision. 
    \item Extensive experiments show that LongSplat achieves state-of-the-art real-time novel view synthesis, providing real-time rendering and reducing Gaussian counts by 44\% compared to existing methods.
\end{itemize}

\begin{figure*}[t]
  \centering
  \includegraphics[width=1.0\linewidth]{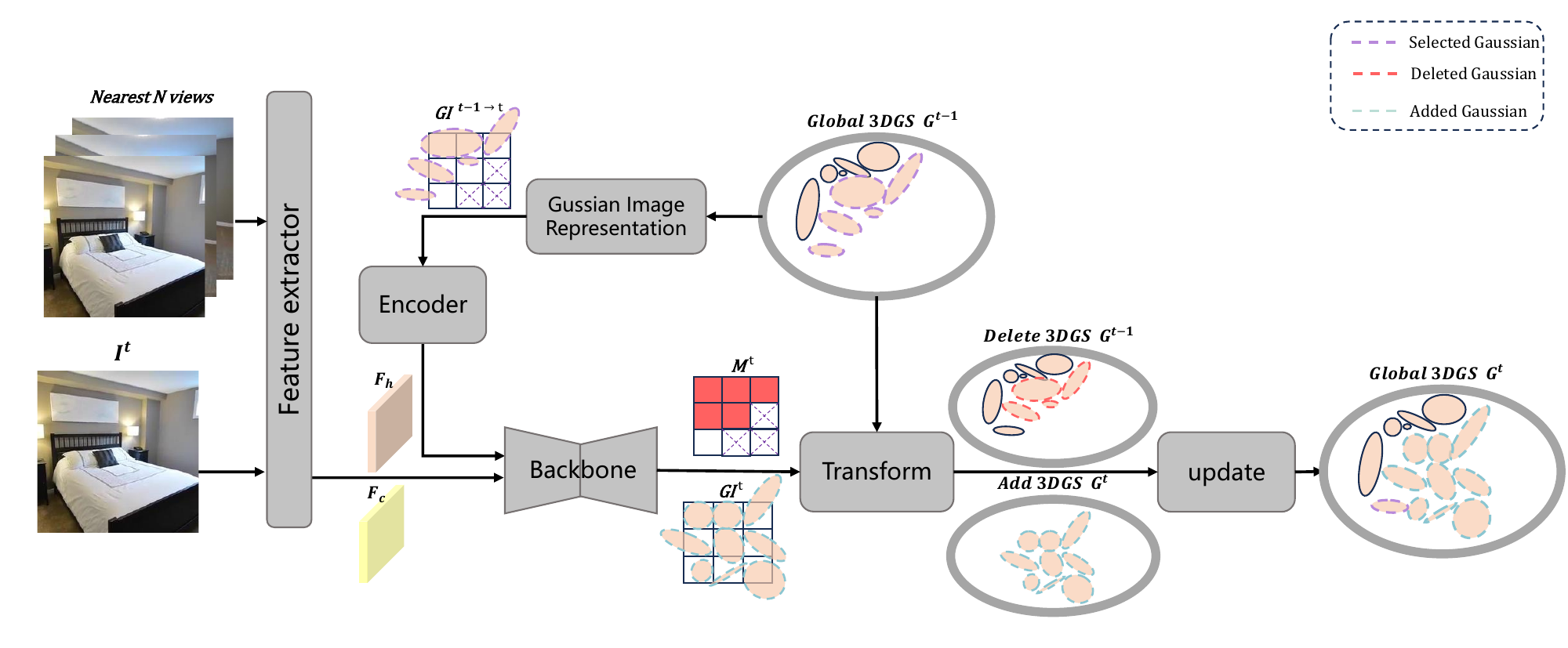}
  \caption{\textbf{Overview of the Longsplat framework.} 
Given an input image sequence $\{I_t\}_{t=1}^T$, our model incrementally constructs a global 3D Gaussian scene representation $\mathcal{G}^g$ through iterative frame-wise updates.
At each timestep $t$, we extract two complementary feature streams: (1) a multi-view spatial feature map $\mathbf{F}_c$ from the current frame and its temporally adjacent neighbors using the DepthSplat pipeline, providing local geometry and appearance cues; and (2) a historical context feature map $\mathbf{F}_h$ by rendering the accumulated global Gaussians $\mathcal{G}^g_{t-1}$ into a 2D Gaussian-Image Representation (GIR) via differentiable projection. These streams are fused via a transformer-based module to produce a fused representation $\mathbf{F}_f$, from which we derive an adaptive update mask $\hat{M}_t$ and generate current-frame Gaussians $\mathcal{G}_t^c$. The global representation $\mathcal{G}^g$ is then selectively updated, enabling efficient long-sequence reconstruction with spatial-temporal consistency.}
  \label{fig:framework}
\end{figure*}

\section{Related Work}
\noindent\textbf{Traditional 3D Gaussian Splatting.}
Traditional 3D Gaussian Splatting (3DGS) methods \cite{kerbl20233d,yu2024mip,huang20242d,lu2024scaffold, gao2025mani} have emerged as a powerful paradigm for high-fidelity novel view synthesis, leveraging explicit 3D Gaussian primitives to represent scenes. 
Unlike Neural Radiance Fields (NeRFs) \cite{mildenhall2021nerf,barron2021mip,barron2023zip,fridovich2023k,cao2023hexplane,liu2022neural,wang2021ibrnet}, which rely on computationally intensive ray-marching-based volume rendering, 3DGS achieves real-time rendering speeds through tile-based rasterization of differentiable Gaussian primitives. 
These methods optimize Gaussian parameters (e.g., position, scale, rotation, and opacity) through per-scene optimization, resulting in high-quality novel view synthesis and fast rendering.
However, the per-scene optimization process is inherently time-consuming, which significantly limits its applicability in real-time perception tasks.

\noindent\textbf{Generalizable 3D Gaussian Splatting.}
Inspired by prior progress in generalizable NeRFs \cite{yu2021pixelnerf,lin2022efficient,gao2022mps,chen2021mvsnerf,ye2023featurenerf}, recent efforts have focused on generalizable 3D Gaussian Splatting methods that enable feed-forward prediction of 3D Gaussians from input images\cite{roh2024catsplat,sheng2025spatialsplat,nam2025generative,bao2024distractor,yan2025instant,kang2025selfsplat,chen2025splatter,tang2024hisplat,zhang2025transplat,wewer2024latentsplat}.
Pioneering works such as PixelSplat \cite{charatan2024pixelsplat} and GPS-Gaussian \cite{zheng2024gps} explore feed-forward 3D Gaussian reconstruction using epipolar geometry from just two input views, achieving fast and high-quality novel view synthesis.
MVS-based methods \cite{chen2024mvsplat, xu2024depthsplat,li2025streamgs} extends this direction by leveraging multi-view geometry through cost volumes for enhanced accuracy and generalization.
Adaptive Gaussian \cite{adaptivegaussianadaptivegaussian} departs from fixed pixel-wise Gaussian representations by dynamically adapting the distribution and number of 3D Gaussians based on local geometric complexity. 

Other approaches extend these approaches to sequential inputs: 
For example, FreeSplat \cite{wang2024freesplat,wang2025freesplat++}  proposes a cross-view aggregation scheme and a pixel-wise triplet fusion strategy that jointly optimizes overlapping view regions, enabling free-view synthesis with geometrically consistent scene reconstruction. 
Yet, due to its latent GS representation, the heavy computational overhead limits scalability to long-sequence inputs, making it less suitable for real-time processing of long-sequence inputs. 
Long-LRM \cite{ziwen2024long} leverages a hybrid architecture merging and Gaussian pruning—to process up to 32 views in a single feed-forward pass, reconstructing entire scenes with performance comparable to optimization-based methods. 
Despite these advances, its reliance on fixed-length reconstruction limits flexibility, making it unsuitable for dynamic, open-ended sequences.
Zpressor~\cite{wang2025zpressor} significantly reduces memory requirements via anchor-frame propagation while achieving high reconstruction quality. 
Compared to such anchor-frame methods that still rely on per-frame prediction and fixed-feature transfer, our approach supports dynamic updates to Gaussians across frames, enabling better use of historical information.
As a per-pixel predictor, it is also compatible with our framework and can serve as a feature encoder within our fusion pipeline.
We propose LongSplat, an online 3D Gaussian reconstruction framework specifically designed for long-sequence inputs, supporting scalable temporal modeling under streaming and interactive conditions. Our approach enables real-time editing and streaming integration without compromising reconstruction fidelity through 3DGS updating and history view fusion techniques.


\noindent\textbf{Indoor Scene Reconstruction}
Indoor scene reconstruction has been extensively studied through various paradigms, including voxel-based methods \cite{sun2021neuralrecon,peng2020convolutional}, TSDF fusion \cite{sayed2022simplerecon}, and Nerf-based approaches \cite{zhu2022nice,zhu2024nicer,rosinol2023nerf,yang2024slam}. 
While these methods excel in geometric reconstruction, they often lack the ability to perform photorealistic novel view synthesis. 
Recent advancements \cite{matsuki2024gaussian,sun2024mm3dgs,ji2024neds,ha2024rgbd,pak2025vigs,xin2025large} introduce 3D Gaussian-based representations that improve rendering speed and quality by avoiding unnecessary spatial computations, but still rely on external or ground-truth depth maps.
In contrast, our method operates end-to-end without depth supervision, leveraging photometric losses to achieve accurate 3D Gaussian localization and scalable scene reconstruction.

\section{Method}
\label{sec:method}

\subsection{Vanilla 3D Gaussian Splatting}
\label{sec:preliminary}
The vanilla 3D Gaussian Splatting (3DGS) represents scenes as a collection of anisotropic Gaussians $\mathcal{G}=\{\mu,\Sigma,c,\alpha\}$, where $\mu$ denotes position, $\Sigma$ the covariance matrix, $c$ the color, and $\alpha$ the opacity. The rendering process follows alpha compositing along each ray: 
\begin{equation}
C(u,v) = \sum_{i\in\mathcal{S}}c_i\alpha_i\prod_{j=1}^{i-1}(1-\alpha_j)
\end{equation}
where $\mathcal{S}$ represents the set of Gaussians sorted by depth.  The Gaussian parameters are optimized by a photometric loss to minimize the difference between renderings and image observations.

\subsection{Longsplat Pipeline}
\label{sec:pipeline}

Longsplat processes an input image sequence ${I}_{t=1}^T$ and iteratively updates the global Gaussian representation $\mathcal{G}^g$. At each timestep $t$, the model takes in the current frame $I_t$ and jointly leverages both multi-view spatial features $\mathbf{F}_c$ and historical global context features $\mathbf{F}_h$ to produce the current per-frame Gaussian predictions $\mathcal{G}^c_t$ and update the global scene representation $\mathcal{G}^g_t$ accordingly.
This process involves extracting two key feature streams:

(1) Multi-view Spatial Feature Map $\mathbf{F}_c$.
To ensure geometric consistency and accurate depth-aware representation, we extract multi-view features from the current frame and its $N$ temporally adjacent neighbors. Specifically, we adopt the feature extraction pipeline from Depthsplat\cite{xu2024depthsplat}, which produces both dense feature maps and per-pixel raw Gaussian predictions. These features capture the local 3D structure and provide a strong prior for the current frame's geometry.

(2) Historical Feature Map $\mathbf{F}_h$.
To incorporate long-range temporal information, we introduce a Gaussian-Image Representation (GIR) that efficiently encodes accumulated global Gaussians $\mathcal{G}^g_{t-1}$ into the current camera view. 
Using a differentiable perspective projection operator $\Pi$, we render the historical Gaussians into a structured, image-aligned 2D format, denoted as $\hat{\mathcal{GI}}^g_t = \Pi(\mathcal{G}^g_{t-1}, K_t, T_t)$, where $K_t$ and $T_t$ are the camera intrinsics and extrinsics. 
The projected Gaussian-Image is then encoded via a shallow CNN to yield the historical context feature map $\mathbf{F}_h$.
\paragraph{History Fusion.}
The spatially aligned features $\mathbf{F}_c$ and temporally accumulated features $\mathbf{F}_h$ are fused through a transformer-based module that attends across both streams to produce an enriched representation $\mathbf{F}_f$. 
This fused feature encodes both current appearance and long-term context. 
From $\mathbf{F}_f$, the module predicts an update mask weighting $\hat{M}_t \in [0, 1]^{H \times W}$, that determines the compression weighting for each pixel.

\paragraph{Compressed Module.}
To determine which Gaussians should be retained or delete, we apply a thresholding strategy to the predicted soft update weights $\hat{M}_t$, producing a binary confidence mask $\mathbf{M}_t \in \{0,1\}^{H \times W}$ based on a tunable confidence threshold. 
This binary mask identifies high-confidence pixels suitable for global storage and further optimization. 
We then use $\mathbf{M}_t$ to filter both the enriched feature $\mathbf{F}_f$ and its corresponding per-pixel learnable embeddings, effectively compressing the Gaussian splatting quantities by discarding uncertain or redundant Gaussians. 
The selected features and embeddings are fed into a lightweight transformer and a shared Gaussian head to produce the final set of compressed per-frame Gaussians $\mathcal{G}^c_t$. 
These are then flagged as valid candidates for global scene representation $\mathcal{G}^g_t$, enabling consistent, efficient, and scalable scene reconstruction over long sequences.

\begin{figure}[t]
  \centering
  \includegraphics[width=1.0\linewidth]{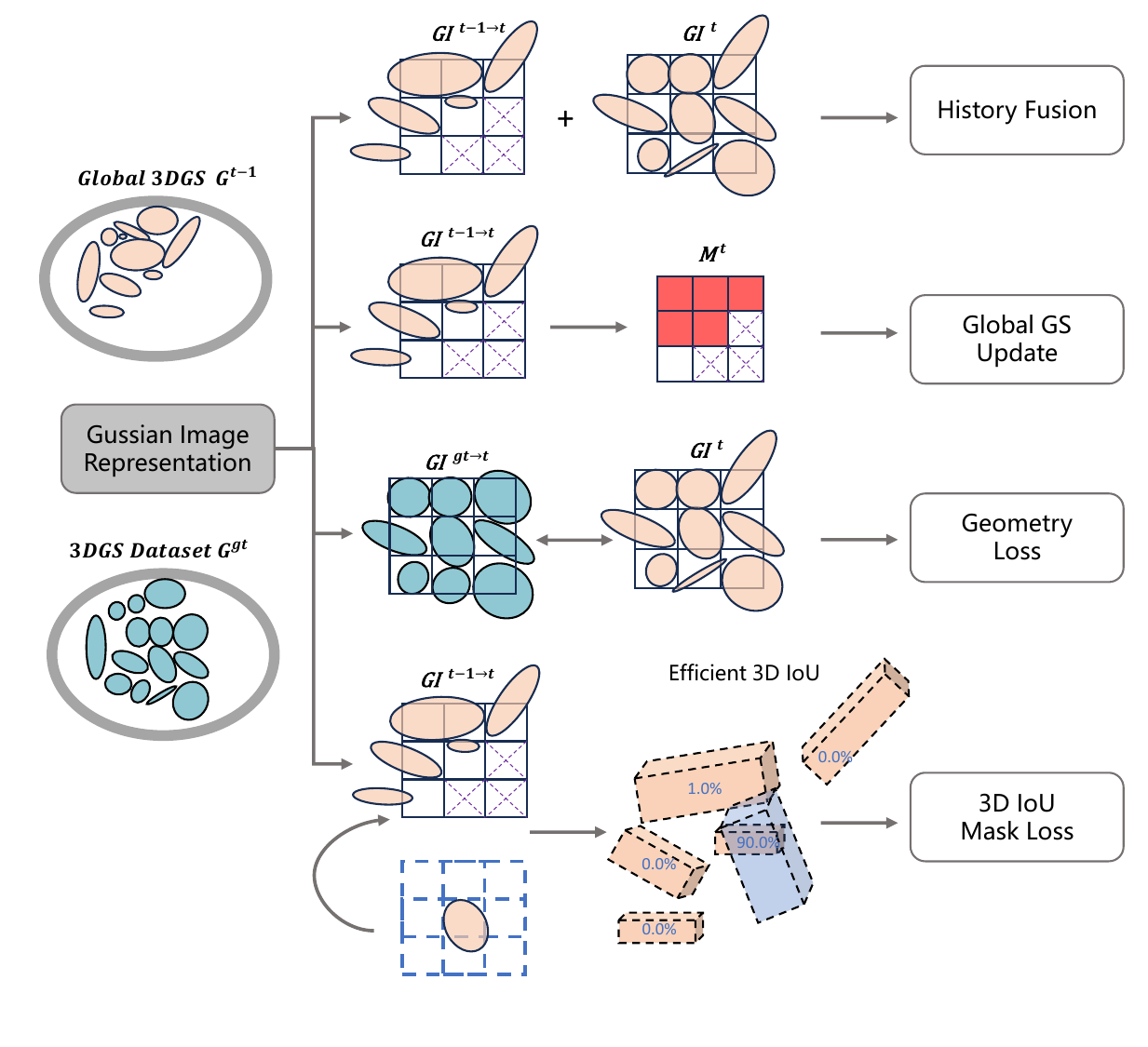}
  \caption{
Overview of the proposed \textbf{Gaussian-Image Representation (GIR)} and its four core capabilities.
GIR encodes per-pixel Gaussian parameters into a structured 2D image space, enabling efficient and flexible 3D reasoning.
\textbf{(a)} \textit{History Fusion:} GIR allows temporally consistent fusion of 3D Gaussians across multiple frames by leveraging shared Gaussian IDs.
\textbf{(b)} \textit{Global Update:} GIR supports gradient-based updates to the global 3D Gaussian field from localized image-space errors.
\textbf{(c)} \textit{Geometry Supervision:} GIR enables pixel-wise geometry loss against ground-truth 3D Gaussians, providing strong spatial supervision.
\textbf{(d)} \textit{Efficient 3D IoU:} By tracking 3D instance masks via Gaussian IDs, GIR enables differentiable 3D IoU estimation and 3D IoU-based mask loss.
}

  \label{fig:introduction}
\end{figure}

\subsection{Gaussian-Image Representation}

\label{sec:gaussian_image}

We propose a \textit{Gaussian-Image Representation} (GIR) that encodes per-pixel Gaussian attributes into a structured 2D format. This compact view-aligned representation enables efficient memory usage, supports localized updates, and bridges the gap between 3D scene modeling and 2D image-space supervision. 

Formally, for each pixel $(u, v)$ in a rendered view, the Gaussian-Image $\mathbf{G}_v \in \mathbb{R}^{H \times W \times 10}$ stores the projected 2D position $\boldsymbol{\mu}^{uv}$, the upper-triangular components of the covariance matrix $\text{vech}(\boldsymbol{\Sigma}^{uv})$, opacity $\alpha^{uv}$, and a unique Gaussian identifier $ID^{uv}$:
\begin{equation}
\mathbf{G}_v(u, v) = \left[\, \boldsymbol{\mu}^{uv},\ \text{vech}(\boldsymbol{\Sigma}^{uv}),\ \alpha^{uv},\ ID^{uv} \,\right]
\end{equation}

Unlike standard 3D Gaussian Splatting (3DGS), which blends all overlapping Gaussians along a ray, our GIR adopts a \textit{sparse rendering strategy} in which each pixel is associated with only a single dominant Gaussian. We consider two selection methods:

\textbf{(1) Nearest Rendering.} The first visible Gaussian (with opacity above threshold $\tau$) is selected:
\begin{equation}
\mathcal{A}(r) = \mathcal{G}_k \quad \text{where} \quad k = \min \left\{ i\ |\ \alpha_i > \tau \right\}
\end{equation}

\textbf{(2) Most-Contributive Rendering.} The Gaussian that contributes most to the final color along the ray is chosen, based on transmittance-weighted opacity:
\begin{equation}
k = \arg\max_i \left( \alpha_i \prod_{j=1}^{i-1}(1 - \alpha_j) \right)
\end{equation}

These rendering strategies produce a clean and disentangled projection of the 3D Gaussians into 2D space, avoiding transparency-induced blurring and enabling more stable per-pixel learning.
To enable consistent mapping between image pixels and their 3D counterparts, we also generate a Gaussian ID map $ID \in \mathbb{Z}^{H \times W}$, recording the index of the selected Gaussian at each pixel. This ID map acts as a lightweight and deterministic link for downstream tasks such as pruning, compression, and temporal fusion.

Moreover, GIR facilitates the creation of training data via a self-supervised bootstrapping process. By first running per-scene optimization (e.g., via LightGaussian~\cite{fan2024lightgaussian}) to obtain high-fidelity 3D Gaussians, we render their corresponding Most-Contributive GIRs as supervision targets. These rendered representations serve as ground truth for learning to predict Gaussian parameters directly from new views, enabling the construction of scalable training datasets without external annotations.

Through this design, GIR provides a unified interface for both training and inference: it enables 2D convolution-based processing of 3D attributes, supports efficient view-wise updates, and grounds learning in a consistent, differentiable projection of the 3D scene.

\subsection{Training}
\label{sec:training}

Our training framework supervises predicted Gaussians using both per-view compressed parameters from LightGaussian~\cite{fan2024lightgaussian} and rendered RGB images, ensuring consistency in geometry, appearance, and compactness. The training loss integrates 3D alignment, mask-guided opacity modulation, and photometric supervision across selected target views.

\textbf{Image Reconstruction Loss.}
To provide dense supervision, we randomly sample a set of target views $\{I_t\}$ from the input sequence and render the current set of predicted Gaussians $\mathcal{G}_t$ into corresponding RGB images $\hat{I}_t$. We then compute a photometric loss between the rendered image and the ground truth:
\begin{equation}
\mathcal{L}_{\text{rgb}} = \sum_{t \in \mathcal{T}} \| \hat{I}_t - I_t \|_1
\end{equation}
This loss provides direct gradients for updating Gaussian attributes, such as color, position, and opacity, and ensures consistency between the 3D representation and the input imagery. Moreover, it enables self-supervised refinement of the predicted confidence mask via opacity modulation (described below).

\textbf{Geometric Alignment.}
In addition to image-level supervision, we enforce consistency between the predicted Gaussians and the ground-truth compressed set $\{\mathcal{G}_v^{gt}\}$. The 3D position alignment loss minimizes spatial deviation:
\begin{equation}
\mathcal{L}_{\text{xyz}} = \frac{1}{|\mathcal{V}|}\sum_{v\in\mathcal{V}}\|\boldsymbol{\mu}_v^{pred} - \boldsymbol{\mu}_v^{gt}\|_1
\end{equation}
The covariance consistency loss promotes shape alignment:
\begin{equation}
\mathcal{L}_{\Sigma} = \frac{1}{|\mathcal{V}|}\sum_{v\in\mathcal{V}}\|\boldsymbol{\Sigma}_v^{pred} - \boldsymbol{\Sigma}_v^{gt}\|_1
\end{equation}
Together, they form the image-space geometric loss:
\begin{equation}
\mathcal{L}_{\text{geo}} = \mathcal{L}_{\text{xyz}} + \lambda_\Sigma \mathcal{L}_{\Sigma}
\end{equation}
where $\lambda_\Sigma = 0.5$.

\textbf{Mask-Guided Learning.}
To enable soft pruning of redundant or outdated Gaussians, we introduce a learnable visibility mask $\mathbf{M}_t \in [0,1]^{H \times W}$, which modulates the rendered opacity:
\begin{equation}
\alpha^{uv}_{\text{mod}} = \mathbf{M}_t(u,v) \cdot \alpha^{uv}
\end{equation}
This mechanism allows the model to compress unnecessary Gaussians while retaining informative ones, guided by both geometric overlap and photometric supervision.

To supervise the mask, we compute a pairwise 3D overlap score based on Oriented Bounding Boxes (OBBs) - the minimal rectangular boxes enclosing each 3D Gaussian ellipsoid. 
Unlike conventional symmetric IoU, we define a view-asymmetric overlap metric:
\begin{equation}
\text{IoU}_p = \max_{q \in \mathcal{N}} \frac{|\text{OBB}_p \cap \text{OBB}_q|}{|\text{OBB}_p|}
\end{equation}
This formulation penalizes historical Gaussians whose OBBs unnecessarily cover fine-scale details in the current view. 
In other words, it encourages the model to mask out large, outdated Gaussians when they are locally replaceable by smaller, view-specific ones.

Enabled by the Gaussian-Image Representation (GIR), these 3D geometric comparisons are efficiently reformulated as local, grid-aligned operations.
Each Gaussian only needs to check collisions with neighbors in a fixed spatial window, allowing the entire process to be implemented as parallel, pixel-wise computations on the GPU—without global scene traversal.
This localized design ensures both computational efficiency and precise redundancy estimation during training.

\begin{figure*}[t]
  \centering
  \includegraphics[width=1.05\linewidth]{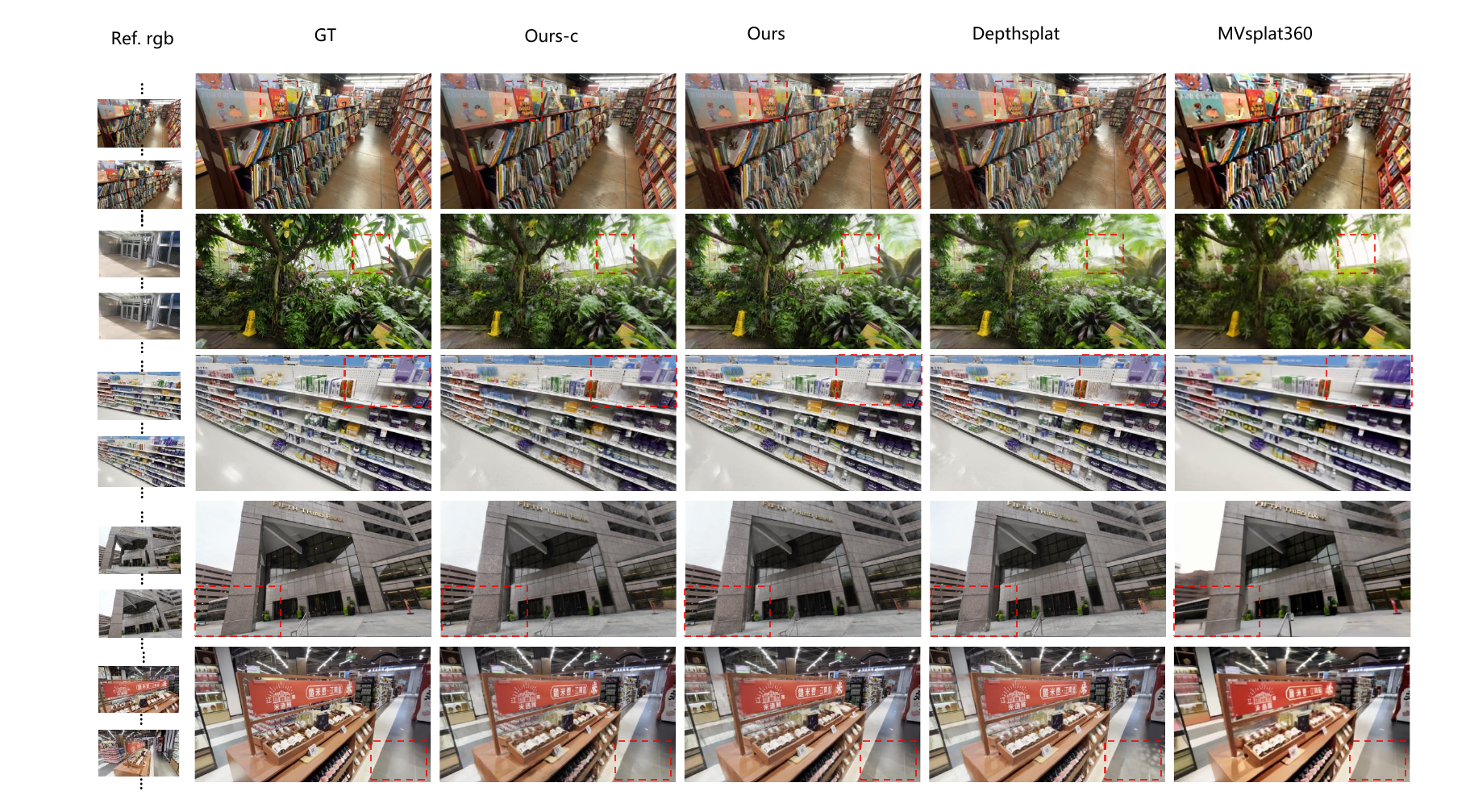}
  \caption{Novel view synthesis on 12 context views. Ref, GT are the input image and the ground truth. Ours-c delete the mask regions. Ours, MVsplat360, and Depthsplat are full-scene results. Our method better removes floaters and preserves fine details, producing more accurate and consistent renderings. )}
  \label{fig:DL3DV}
\end{figure*}

To supervise $\mathbf{M}_t$, we treat Gaussians with high overlap as redundant and assign ground truth $\mathbf{M}_t^{gt}=1$, and others with $\mathbf{M}_t^{gt}=0$. We then apply a weighted binary cross-entropy loss:
\begin{equation}
\mathcal{L}_{\text{mask}} = \frac{1}{|\Omega|} \sum_{(u,v) \in \Omega} \lambda\left(u,v\right) \cdot \text{BCE}\left(\mathbf{M}_t(u,v),\ \mathbf{M}_t^{gt}(u,v)\right)
\end{equation}
where $\Omega$ is the set of all pixel locations, and the weight function is defined as:
\begin{equation}
\lambda(\mathbf{M}_t^{gt}) = 
\begin{cases}
\lambda_{\text{pos}}, & \text{if } \mathbf{M}_t^{gt}\left(u,v\right) = 1 \\
\lambda_{\text{neg}}, & \text{if } \mathbf{M}_t^{gt}\left(u,v\right) = 0
\end{cases}
\end{equation}

We set $\lambda_{\text{pos}} > \lambda_{\text{neg}}$ to counteract the photometric loss, which implicitly encourages maintaining opacity. Our mask loss explicitly penalizes geometric redundancy, ensuring better spatial compactness and consistency across frames.

\textbf{Total Objective.}
The final loss combines all components into a multi-task objective:
\begin{equation}
\mathcal{L}_{\text{total}} = \mathcal{L}_{\text{rgb}} + \mathcal{L}_{\text{geo}} + \mathcal{L}_{\text{mask}} .
\end{equation} 

\section{Experiments}
\label{sec:experiment}

\textbf{Implementation Details}
We adopt DepthSplat\cite{xu2024depthsplat} as our baseline framework while keeping all its parameters fixed during training. 
The feature representations and Gaussian splatting outputs from DepthSplat are directly utilized as our model inputs. 
During rendering, we consistently use 10 target views for loss computation, ensuring multi-view consistency in the optimization process. 
For optimization, we employ the AdamW optimizer with a base learning rate of $1\times10^{-4}$.
During training, the number of input views is randomly sampled between 2 and 8. The model is trained with a resolution downsampling factor of 1/8 (i.e., 256$\times$448 resolution).  For hardware configuration, we employ 8 $\times$ RTX 4090 GPUs to perform 10K base training iterations for the uncompressed model, followed by an additional 10K iterations on 4 $\times$ H100 GPUs for the compressed version.


\textbf{Training Datasets} 
We conduct training on the DL3DV-10K\cite{ling2024dl3dv} dataset, which consists of 9,896 training scenes and 140 official test scenes. Additionally, we provide optional auxiliary data reconstructed through DepthSplat (24-view inputs) with LightGaussian compression, which contributes to conditional auxiliary losses when available during training.
We selected 6,845 scenes meeting our quality criteria: compression rate $>$30\% and PSNR $>$28.0. These selected scenes achieve an average PSNR of 31.50 with 69.60\% average compression rate.

\subsection{DL3DV Benchmark Evaluation}


\begin{table}[h]
\small
\centering
\caption{Multi-view comparison with state-of-the-art methods}
\label{tab:DL3DV}
\begin{tabular}{@{}l@{\hspace{0.8em}}l@{\hspace{0.8em}}c@{\hspace{0.8em}}c@{\hspace{0.8em}}c@{\hspace{1em}}c@{}}
\toprule
\textbf{Views} & \textbf{Method} & \textbf{PSNR↑} & \textbf{SSIM↑} & \textbf{LPIPS↓} & \textbf{c-ratio↑} \\
\midrule
\multirow{3}{*}{12} 
& MVSplat-360 & 17.05  & 0.4954 & 0.3575 &  0.00\% \\
& DepthSplat & 22.02 & 0.7609 & 0.2060 & 0.00\% \\
& Ours & 22.68 & 0.7824 & 0.1923 & 0.00\% \\
& Ours-c & 21.69 & 0.7482 & 0.2213 & 25.52\% \\
\hline
\multirow{3}{*}{50}
& MVSplat-360 & OOM & / &  / & / \\
& DepthSplat & 21.39 & 0.7341 &  0.2212 &  0.00\% \\
& Ours & 23.71 & 0.8159 & 0.1683 &   0.00\% \\
& Ours-c & 23.54 & 0.8056 &  0.1742 & 43.77\% \\
\hline
\multirow{3}{*}{120}
& DepthSplat & 17.77 & 0.5899 & 0.3622  &  0.00\% \\
& Ours &  21.02 & 0.7176 & 0.2608  & 0.00\% \\
& Ours-c & 21.34 & 0.7345  & 0.2449 & 44.37\% \\
\bottomrule
\end{tabular}
\vspace{2mm}
\parbox{\linewidth}{\scriptsize
\textbf{Abbreviations}: "c-ratio" shows the percentage of compressed Gaussians. "Ours-c" is our compressed strategy applied.}
\end{table}

\vspace{-20pt}
\paragraph{Quantitative Results.}
As shown in Table~\ref{tab:DL3DV}, our method consistently outperforms DepthSplat across all view counts, achieving state-of-the-art results in both quality and compactness. We evaluate under 12, 50, and 120 context-view settings, corresponding to 100, 50, and 120 fixed target views, respectively. For large-view settings (50 and 120 views), we adopt DepthSplat’s MVS-based feature extraction with temporal sampling at fixed 10-frame intervals to ensure stable inputs. Our method then performs sequential 3D Gaussian reconstruction guided by these features.

For evaluation, we test under 12, 50, and 120-view configurations with fixed target sets to ensure fairness and scalability.
Under sparse 12-view input, our full model achieves 22.68~PSNR, outperforming DepthSplat (22.02) with improved SSIM and LPIPS. Our compressed variant (Ours-c) retains competitive quality (21.69 PSNR) while reducing Gaussian count by 25.52\%, demonstrating effective compactness.
As view count increases, DepthSplat's performance degrades due to uncontrolled Gaussian growth (17.77 PSNR at 120 views), while our approach remains stable.
At 50 views, our full model reaches 23.71~PSNR (+2.32~dB over DepthSplat), and Ours-c achieves 23.54 while removing 43.77\% of Gaussians. At 120 views, Ours-c still maintains high fidelity (21.34 PSNR, 0.2449 LPIPS) with 44.37\% compression.
These results highlight our method’s scalability and efficient long-sequence reconstruction.

\paragraph{Qualitative Results.} 
Figure~\ref{fig:DL3DV} shows qualitative results on several DL3DV scenes with 12 input views. We compare our method (Ours), its compressed variant (Ours-c), MVsplat360\cite{chen2024mvsplat360}, and our baselines, DepthSplat\cite{xu2024depthsplat}.
DepthSplat tends to produce floaters and blurred surfaces in complex regions, while MVsplat360 occasionally exhibits structural inconsistencies when applied to long sequences.
Our method generally produces sharper and more consistent reconstructions, with clearer geometry and fewer visual artifacts.
Notably, the compressed variant (Ours-c) maintains comparable quality, and in some cases—such as the first row—shows improved clarity in fine details like book spines. This suggests that removing low-confidence Gaussians may help reduce visual clutter and enhance overall fidelity.


\subsection{Ablation and Analysis}


\begin{figure*}[t]
  \centering
  \includegraphics[width=1.0\linewidth]{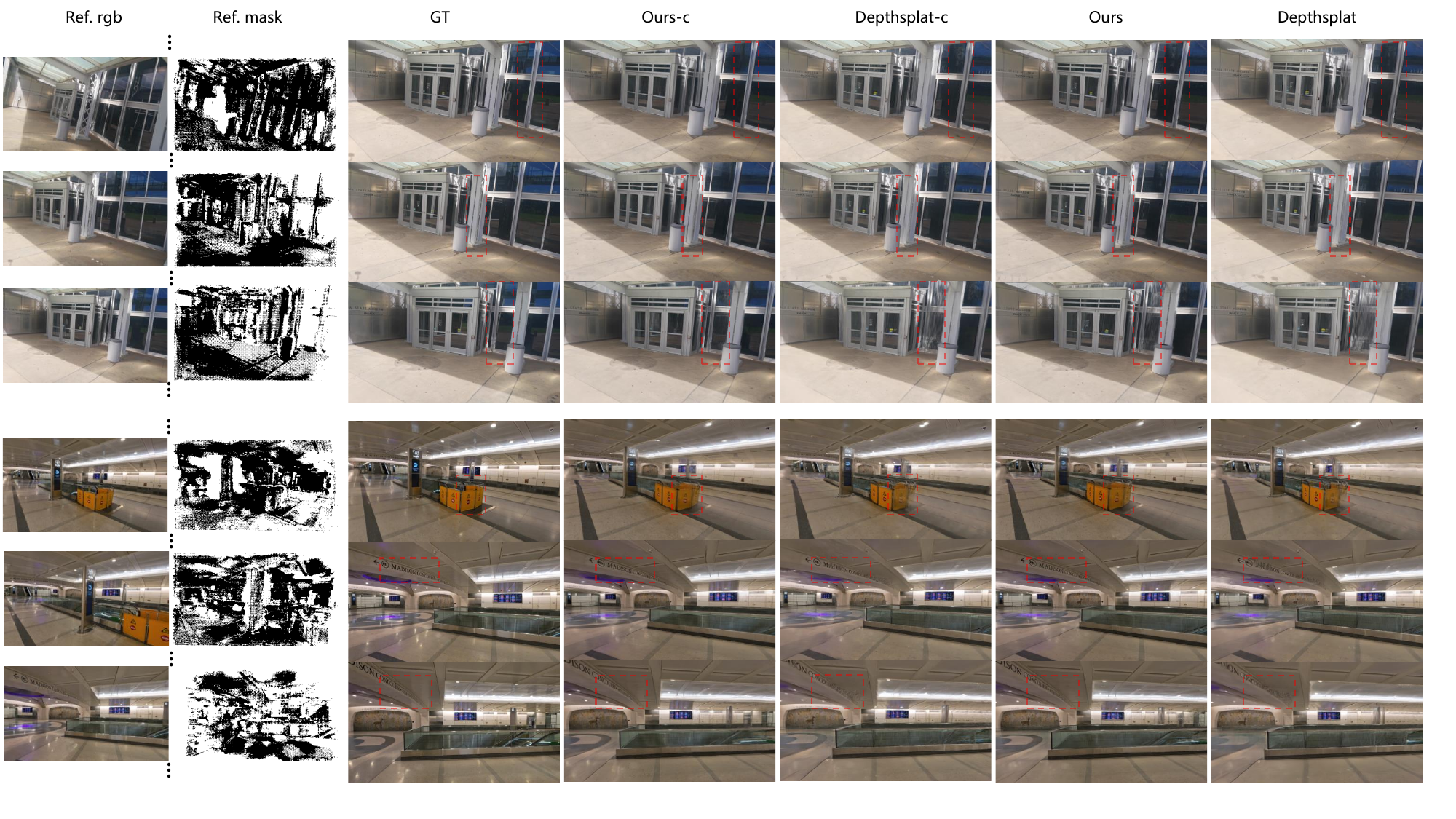}
  \caption{Novel view synthesis on 50 context views. Ref. rgb and Ref. mask are the input image and compression mask. GT is the ground truth. Ours-c and Depthsplat-c delete the mask regions. Ours and Depthsplat are full-scene results. Our method better removes floaters and preserves fine details, producing more accurate and consistent renderings.)}
  \label{fig:ablation}
\end{figure*}

\paragraph{Ablation Study on Component-wise Contributions.}
Table~\ref{tab:ablation} reports the incremental impact of each core module. 
Adding Unet refinement (\textbf{U}) to the baseline improves PSNR from 21.39dB to 21.71dB. 
History fusion (\textbf{F}) brings a larger boost to 22.78dB PSNR and significantly lowers LPIPS, highlighting the value of temporal context.
Introducing 3D supervision (\textbf{D}) further enhances alignment, reaching 23.12 PSNR.
We then examine compression and masking. Applying a fixed compression mask on the 3D dataset (+D\&M) reduces Gaussians by 43.77\% but lowers PSNR to 22.47dB, indicating that static pruning can discard relevant points. Introducing the learned compression module (\textbf{C}) without masking restores high fidelity (23.71dB, SSIM 0.8159, LPIPS 0.1683) while keeping c‑ratio at 0\%. Finally, combining compression with adaptive masking (+C\&M) retains most of this gain (PSNR 23.54dB, SSIM 0.8056, LPIPS 0.1742) at a 43.77\% compression ratio. 
These results validate that each component contributes complementary gains, with joint compression and masking offering the best balance of quality and compactness.

\begin{table}[t]
\centering
\scriptsize
\caption{Ablation study on core modules}
\label{tab:ablation}
\begin{tabular}{@{}l@{\hspace{1mm}}ccccc@{\hspace{1mm}}cccc@{\hspace{1mm}}}
\toprule
\multirow{2}{*}{\textbf{Config}} & 
\multicolumn{5}{c@{\hspace{1mm}}}{\textbf{Components}} & 
\multicolumn{4}{c}{\textbf{Metrics}} \\
\cmidrule(lr){2-6} \cmidrule(l){7-10} 
 & U & F & D & C & M & PSNR$\uparrow$ & SSIM$\uparrow$ & LPIPS$\downarrow$ & c-ratio$\uparrow$ \\
\midrule
Baseline      & -- & -- & -- & -- & -- & 21.39 & 0.7341  & 0.2212 & 0.00 \\
+ Unet        & \checkmark & -- & -- & -- & -- & 21.71  & 0.7463 & 0.2172 & 0.00 \\
+ Fusion      & \checkmark & \checkmark & -- & -- & -- & 22.78 & 0.7934 & 0.1777 & 0.00 \\
+ 3D Dataset  & \checkmark & \checkmark & \checkmark & -- & -- & 23.12 & 0.8087 & 0.1679 & 0.00 \\
+ 3D Dataset(M)  & \checkmark & \checkmark & \checkmark & -- & \checkmark & 22.47 & 0.7793 & 0.1851 & 43.77 \\
+ Compression       & \checkmark & \checkmark & \checkmark & \checkmark & -- & 23.71 & 0.8159  & 0.1683 & 0.00 \\
+ Compression(M)   & \checkmark & \checkmark & \checkmark & \checkmark & \checkmark &  23.54 & 0.8056 & 0.1742 & 43.77 \\
\bottomrule
\end{tabular}
\vspace{2mm}
\parbox{\linewidth}{\scriptsize
\textbf{Abbreviations:} \texttt{U}: Unet, \texttt{F}: History fusion, \texttt{D}: 3D dataset, 
\texttt{C}: Compress module, \texttt{M}: Compress mask.
\textbf{Note:} Context-view and target-view sequences are both 50 frames.}
\vspace{-20pt}
\end{table}


\begin{table}[h]
\centering
\small
\caption{Performance under varying masking thresholds $\tau$.}
\label{tab:confidence_comparison}
\begin{tabular}{@{}llcccc@{}}
\toprule
\textbf{$\tau$} & \textbf{Method} & \textbf{PSNR} & \textbf{SSIM} & \textbf{LPIPS$\downarrow$} & \textbf{c-ratio.$\uparrow$} \\
\midrule
\multirow{2}{*}{No Mask} 
  & DepthSplat & 21.39 & 0.7341 & 0.2212 & 0.0\% \\
  & Ours       & \textbf{23.71} & \textbf{0.8159} & \textbf{0.1683} & 0.0\% \\
\midrule
\multirow{2}{*}{0.1} 
  & DepthSplat & 21.82 & 0.7497 & 0.2081 & 23.32\% \\
  & Ours       & \textbf{23.66} & \textbf{0.8141} & \textbf{0.1680} & 23.32\% \\
\midrule
\multirow{2}{*}{0.3} 
  & DepthSplat & 22.17 & 0.7621 & 0.1996 & 38.11\% \\
  & Ours       & \textbf{23.61} & \textbf{0.8092} & \textbf{0.1716} & 38.11\% \\
\midrule
\multirow{2}{*}{0.5} 
  & DepthSplat & 22.32 & 0.7668 & 0.1964 & 43.77\% \\
  & Ours       & \textbf{23.54} & \textbf{0.8056} & \textbf{0.1742} & 43.77\% \\
\bottomrule
\end{tabular}
\end{table}

\vspace{-10pt}
\paragraph{Evaluation of Masking Strategies Under Confidence Thresholds.}

We evaluate the effect of varying the masking threshold $\tau$ on reconstruction quality and compression (Table~\ref{tab:confidence_comparison}). As $\tau$ increases, DepthSplat exhibits gradual improvement in PSNR and LPIPS. However, this gain does not stem from better modeling but rather from aggressive pruning of spatially inconsistent or floating points, which often produce visible artifacts. These improvements are thus partially due to the suppression of such artifacts rather than true reconstruction fidelity.
Our method leverages a learned confidence-based mask to explicitly identify and remove redundant or noisy Gaussians. As a result, even at high pruning ratios (e.g., $\tau=0.5$, 43.77\% compression), our method maintains high PSNR (23.54~dB) and low LPIPS (0.1742), nearly matching the unpruned case (23.71~dB, 0.1683). This demonstrates that our compression is both precise and structure-aware, effectively reducing redundancy while preserving essential geometry and appearance.
These trends are also visually reflected in Fig.~\ref{fig:ablation}, where our method removes floaters and preserves detail, while DepthSplat tends to oversimplify or introduce inconsistencies in compressed regions.

\section{Conclusion}
We present LongSplat, a real-time 3D Gaussian Splatting framework tailored for long-sequence reconstruction. To address scalability and redundancy issues in existing feed-forward pipelines, LongSplat introduces an incremental update mechanism that compresses redundant Gaussians and incrementally integrates current-view observations into a consistent global scene. Central to our design is the Gaussian-Image Representation (GIR), which projects 3D Gaussians into structured 2D maps for efficient fusion, identity-aware compression, and 2D-based supervision. By enabling lightweight per-frame updates and effective historical modeling, LongSplat mitigates memory overhead and quality degradation in dense-view settings. Extensive experiments show that it achieves real-time rendering, improves visual quality, and reduces Gaussian redundancy by over 44\%, offering a scalable solution for high-quality online 3D reconstruction.

In future work, we aim to incorporate stronger, pose-free extractors such as Cust3r\cite{wang2025continuous}, VGGT\cite{wang2025vggt}, and DUST3R\cite{wang2024dust3r}, removing the need for pre-computed camera poses. We also plan to extend LongSplat with semantic reasoning for 3D understanding in embodied scenarios.

\newpage

{\small
\bibliographystyle{ieee_fullname}
\bibliography{main}
}

\clearpage

\end{document}